%% file: main.tex
\documentclass[10pt,twocolumn,letterpaper]{article}

\usepackage[pagenumbers]{cvpr}

\input{preamble}

\definecolor{cvprblue}{rgb}{0.21,0.49,0.74}
\usepackage[pagebackref,breaklinks,colorlinks,allcolors=cvprblue]{hyperref}
\hypersetup{
  pdftitle={Reg4AD: Learning to Interpret Registration Discrepancies for Multi-Category 3D Anomaly Detection},
  pdfauthor={Xuanming Cao, Hao Xu, Chengyu Tao}
}

\def\confName{CVPR}
\def\confYear{2027}

\definecolor{titleblue}{RGB}{0,92,185} % 
\newcommand{\ours}{\textsc{Reg4AD}}%
\newcommand{\hl}[1]{\textcolor{titleblue}{#1}}

\title{\hl{\ours}: Learning to Interpret \hl{Reg}istration Discrepancies \hl{for} Multi-Category 3D \hl{A}nomaly \hl{D}etection}

\author{%
Xuanming~Cao$^{1}$ \quad Hao~Xu$^{2}$ \quad
Chengyu~Tao$^{3}$\thanks{Corresponding author.}\\
{\small $^{1}$Yunnan Minzu University, Kunming, China}\\
{\small $^{2}$University of Sydney, Sydney, Australia}\\
{\small $^{3}$Hunan University, Changsha, China}\\
{\small\ttfamily xuanming@ymu.edu.cn \quad ggg43127748@gmail.com \quad taochengyu@hnu.edu.cn}
}

\begin{document}
\maketitle
\input{sec/0_abstract}    
\input{sec/1_intro}

\input{sec/2_related}
\input{sec/3_method}
\input{sec/4_expriments}
\input{sec/5_conlusion}

\input{sec/6_llm_usage}
{
    \small
    \bibliographystyle{ieeenat_fullname}
    \bibliography{main}
}

% Include the supplementary material in the preprint.
\input{sec/X_suppl}

\end{document}

%% file: preamble.tex
\usepackage{placeins}
\usepackage{multirow}

%% file: sec/0_abstract.tex
\begin{abstract}
Registration-based 3D anomaly detection aligns test point clouds with normal references to identify anomalies. However, anomaly-induced structural differences can disrupt alignment, and post-registration discrepancies may reflect a combination of anomalies, normal shape variation, and alignment errors. This ambiguity limits the reliability of using geometric distances directly as anomaly scores. To address this problem, we propose \ours, a unified framework that learns to interpret registration discrepancies for multi-category 3D anomaly detection. \ours \ combines local residual context, which captures spatial patterns in geometric residuals, with multi-scale structural discrepancies derived from local descriptors. A context-structure fusion network integrates these cues to predict pointwise anomaly scores, which are then aggregated for object-level detection. Trained with synthetic anomalies generated from normal point clouds, \ours \ uses one shared model across all categories within each dataset. On Real3D-AD, \ours \ achieves mean object-level and point-level AUROC of 92.0\% and 95.6\%, respectively. On Anomaly-ShapeNet, it achieves 95.7\% and 98.2\% across 40 categories. \ours\ achieves state-of-the-art performance on both benchmarks. Ablation studies demonstrate the importance and complementary roles of local residual context and multi-scale structural discrepancies.
\end{abstract}

%% file: sec/1_intro.tex
\section{Introduction}
\label{sec:intro}

Industrial anomaly detection (AD) supports quality control and automated inspection in manufacturing~\cite{cao2026iaenet,bergmann2019mvtec}. Image-based methods often use pretrained visual features to detect anomalies in surface appearance~\cite{roth2022towards,xu2026ano,tao2024f2pad,deng2022anomaly,liu2023simplenet,guo2025dinomaly}. However, shallow dents and protrusions can be difficult to detect in 2D images when they produce little contrast in color or texture. By measuring surface geometry directly, 3D sensors provide point clouds that can capture the depth of dents and the height of protrusions~\cite{bergmann2022mvtec3d}. This capability has contributed to growing interest in 3D AD~\cite{liu2023real3d,li2024towards}.

Given the scarcity and diversity of anomalous samples, industrial 3D AD commonly adopts an unsupervised setting in which models learn from anomaly-free training samples~\cite{bergmann2022mvtec3d,liu2023real3d}. Representative methods characterize normality using feature memories, reconstructed geometry, or implicit surfaces~\cite{liu2023real3d,li2024towards,zheng2025bridging}. Among these approaches, Reg3D-AD~\cite{liu2023real3d}, PASDF~\cite{zheng2025bridging}, and Reg2Inv~\cite{yu2025registration} register a test point cloud to a normal reference and assign pointwise anomaly scores using geometric or feature distances.

\begin{figure}[t]
    \centering
    \includegraphics[width=\columnwidth,pagebox=cropbox]{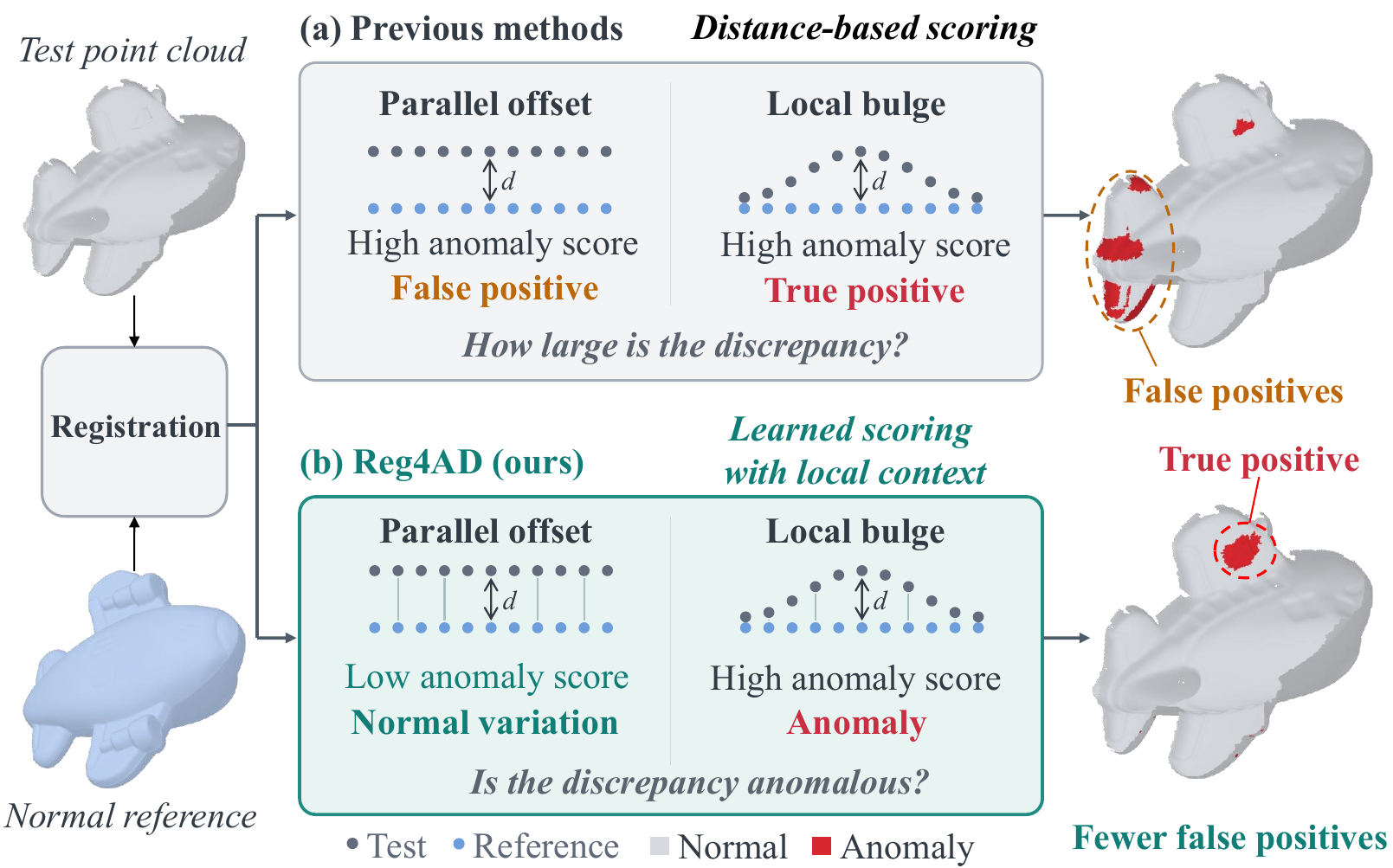}
    \caption{Comparison of registration-based 3D anomaly detection methods. (a) Previous registration-based pipelines score anomalies using predefined geometric or feature distances. The center points of a parallel offset and a local bulge are both at distance $d$ from the reference, so geometric distance assigns them the same high anomaly scores. (b) \ours \ learns to interpret post-registration discrepancies for anomaly scoring. We use neighboring residuals and local shape to distinguish the normal variation from the anomalous bulge. On the Real3D-AD \cite{liu2023real3d} airplane shown on the right, \ours \ localizes the anomaly with fewer false positives.}
    \label{fig:motivation}
\end{figure}

Although registration-based methods are effective, we argue that post-registration discrepancies should not be directly equated with anomaly scores. Rigid registration aligns scans of the same object by estimating their relative pose. In AD, even normal samples from the same production batch can differ in shape, and anomalies introduce further structural changes. These geometric differences can make accurate rigid alignment difficult. The discrepancies remaining after registration may therefore reflect a combination of normal shape variation, anomalous changes, and registration errors. The magnitude of a discrepancy alone is insufficient to determine whether it indicates an anomaly.

Existing registration-based methods that rely solely on discrepancy magnitude can produce false positives in regions with normal shape variation or registration errors, as shown in Fig.~\ref{fig:motivation}(a). Our key insight is that post-registration discrepancies provide useful evidence for AD, but predefined distance measures alone are insufficient to interpret this evidence (Fig.~\ref{fig:motivation}(b)). Therefore, the research problem is how to learn anomaly cues from post-registration discrepancies. We address this question by examining two cues: residual variation across neighboring points and local shape differences between a test point cloud and its normal reference.

Specifically, we propose \ours{}, a unified model for multi-category 3D AD that represents these two cues using \textbf{Local Residual Context (LRC)} and \textbf{Multi-scale Structural Discrepancy Encoding (MSDE)}. The model first registers the input point cloud to a normal reference from the same category. For each aligned point, LRC encodes the residual magnitude together with the mean and standard deviation of neighboring residuals. This context helps the scorer assess whether a large residual is shared by nearby points (Fig.~\ref{fig:motivation}). MSDE extracts multi-scale local shape descriptors from both point clouds using a pretrained encoder and compares descriptors from corresponding patches at each scale. A \textbf{Context-Structure Fusion Network (CSFN)} fuses the two representations to predict pointwise anomaly scores, which are then aggregated into an object-level score.

We jointly train the learnable descriptor projections and CSFN using normal scans and synthetic local geometric anomalies, while keeping the registration module and pretrained encoder frozen. CSFN learns to interpret residual patterns and structural discrepancies jointly when predicting anomaly scores. A single model is shared across all categories within each dataset.

Our contributions are as follows:
\begin{itemize}
    \item We propose \ours, a unified model that learns to interpret registration discrepancies for accurate multi-category 3D AD.
    \item We introduce two complementary representations of post-registration discrepancies. LRC captures how residuals vary across neighboring points, while MSDE encodes local shape differences between the input and its normal reference at multiple scales.
    \item We develop CSFN, which learns to predict an anomaly score for each point by fusing LRC and MSDE representations.
    \item On Real3D-AD and Anomaly-ShapeNet, \ours\ achieves the best performance among the compared methods. These results support the effectiveness of learning to score registration discrepancies.
\end{itemize}

%% file: sec/2_related.tex
\section{Related Work}
\label{sec:related_work}

\subsection{3D Anomaly Detection}
\label{sec:rw_3d_ad}

3D AD is commonly studied in an unsupervised setting, where models learn normal patterns only from anomaly-free samples~\cite{liu2023real3d,li2024towards}. Existing methods differ mainly in how they represent normality and measure deviations from it. Some compare a test sample with normal features: BTF~\cite{horwitz2023back} uses handcrafted geometric descriptors, and CPMF~\cite{cao2024complementary} combines them with features from rendered views. M3DM~\cite{wang2023multimodal}, Shape-Guided~\cite{chu2023shape}, IAENet~\cite{cao2026iaenet}, and G$^2$SF~\cite{tao2025g2sf} further exploit RGB information through feature, memory, or score fusion. Others learn to recover normal geometry and score deviations from it. IMRNet~\cite{li2024towards} iteratively masks and reconstructs point patches, R3D-AD~\cite{zhou2024r3d} learns diffusion-based reconstruction from synthetic anomalies, and PO3AD~\cite{ye2025po3ad} predicts point offsets whose magnitudes serve as anomaly scores. However, these methods maintain a separate model or memory bank per object category, which becomes costly as the category count grows.

\subsection{Multi-category 3D Anomaly Detection}
\label{sec:rw_multiclass_3d_ad}

Multi-category 3D AD trains one model to detect and localize anomalies across object categories~\cite{cheng2025mc3d,liang2026mff}. Such a model must accommodate diverse normal geometries while retaining the local detail needed to identify anomalies. Recent approaches pursue unified reconstruction and memory-based feature matching.
For unified reconstruction, MC3D-AD~\cite{cheng2025mc3d} uses geometry-aware attention, while MFF-M3AD~\cite{liang2026mff} combines multi-scale feature fusion, refinement, and reconstruction. SeDiR~\cite{kim2026semantically} couples category-conditioned contrastive learning with geometry-guided decoding to reduce inter-category entanglement. For feature matching, GLFM~\cite{cheng2025boosting} adapts representations using synthetic anomalies, matching them against global and local normal memories.
In contrast, \ours{} learns category-agnostic anomaly cues from registration residuals and local shape discrepancies, enabling unified 3D AD.

\begin{figure*}[!t]
    \centering
    \includegraphics[width=\textwidth,pagebox=cropbox]{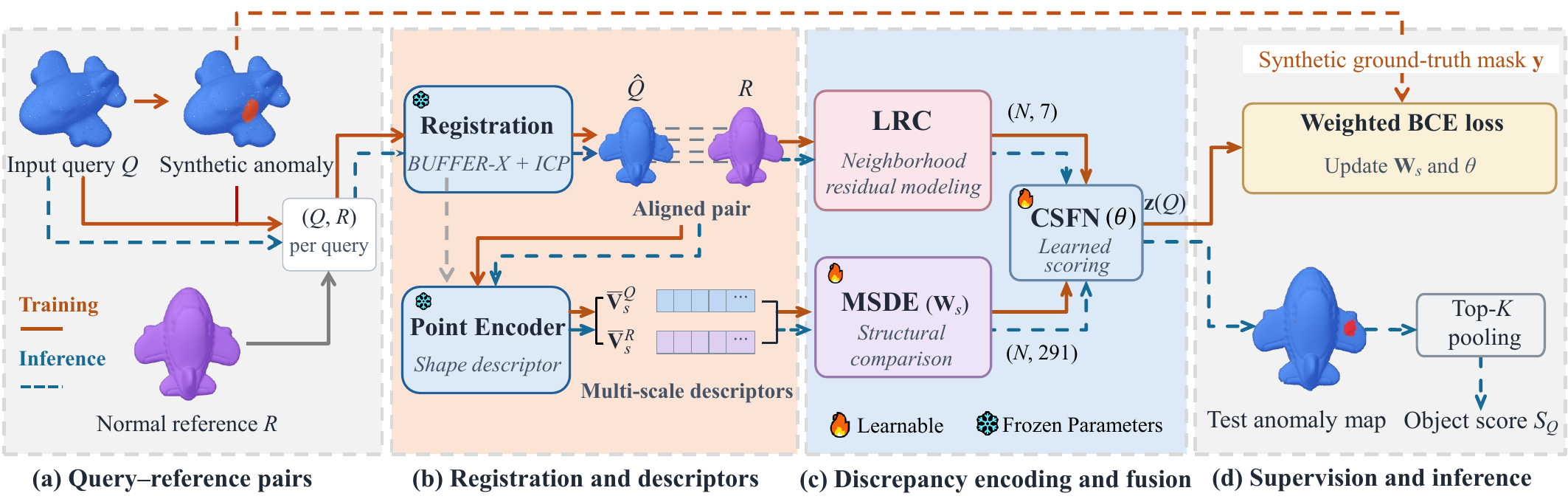}
    \caption{Overview of \ours. (a) Each query is paired with a normal reference from its category. (b) BUFFER-X and ICP align the query to the reference, and a frozen point encoder supplies local descriptors at three scales. (c) CSFN fuses neighborhood residual statistics (LRC) and local shape discrepancies (MSDE) to score each query point. (d) During training (orange), a weighted binary cross-entropy loss uses the synthetic mask to supervise scores on the anomalous query. At inference (blue), averaging the top $K$ scores in the test anomaly map gives the object score.}
    \label{fig:overview}
\end{figure*}

\subsection{Registration-based 3D Anomaly Detection}
\label{sec:rw_registration_ad}

Registration-based methods establish a common geometric frame for comparing test point clouds with normal references, reducing differences caused by object pose. Existing methods differ in how they align inputs and how they compare them with normality. Reg3D-AD~\cite{liu2023real3d} registers each input to a normal reference and scores it with local and global feature memories. PASDF~\cite{zheng2025bridging} aligns inputs to a canonical pose and scores points by their distances to a normal surface represented by a continuous signed distance function. Reg2Inv~\cite{yu2025registration} learns rotation-invariant descriptors through registration training and uses them both for registration and for memory-based scoring. These methods mainly improve how inputs are aligned and represented, while the comparison after alignment remains a predefined geometric or feature distance.

Normal shape variation and anomaly-induced alignment errors make residual magnitude alone unreliable for anomaly scoring. \ours{} combines local residual context with multi-scale structural discrepancies to learn which post-registration deviations indicate anomalies.

%% file: sec/3_method.tex
\section{Method}
\label{sec:method}

\subsection{Problem Definition and Overview}
\label{sec:method_overview}

In unified multi-category 3D AD, a single model is trained on normal point clouds from multiple object categories, with its parameters shared across all categories within a dataset. Let $\mathcal D=\{Q_t\}_{t=1}^{n}$ denote the set of $n$ normal training samples, where $Q_t$ is the $t$-th sample. For an input $Q=\{\mathbf q_i\}_{i=1}^{N}$ with $N$ points $\mathbf q_i\in\mathbb R^3$, the model predicts pointwise anomaly scores $\mathbf z(Q)\in\mathbb R^N$ and an object-level score $S_Q\in\mathbb R$, with higher scores indicating anomalies.

\ours\ aligns each query $Q$ to a normal reference $R$ from the same category and extracts local descriptors at three scales with a frozen encoder (Fig.~\ref{fig:overview}(a)--(b)). LRC summarizes residuals around each aligned point, while MSDE compares descriptors of corresponding patches. The CSFN fuses these cues into a pointwise anomaly map (Fig.~\ref{fig:overview}(c)). We train the MSDE projections and CSFN using point labels from clean and synthetically anomalous scans. At inference, averaging the top $K$ scores gives the object score (Fig.~\ref{fig:overview}(d)).

\subsection{Reference-guided Registration}
\label{sec:registration}

We register $Q$ to its corresponding normal reference $R=\{\mathbf r_j\}_{j=1}^{M}$, using the same normal-reference protocol as PASDF~\cite{zheng2025bridging}. Here, $M$ is the number of reference points and $\mathbf r_j\in\mathbb R^3$ is the $j$-th reference point. After dataset-specific coordinate preprocessing, we use a frozen, pretrained BUFFER-X model~\cite{lim2026toward} to estimate a coarse rigid transformation $T_{\mathrm{coarse}}$ from $Q$ to $R$. We choose BUFFER-X for its ability to robustly align point clouds in unseen scenes without fine-tuning. Point-to-point iterative closest point (ICP) refinement~\cite{besl1992method} then produces an incremental transformation $T_{\mathrm{ICP}}$. The final transformation $T$ maps each query point to its aligned position $\hat{\mathbf q}_i$, forming the aligned point cloud $\hat Q$:
\begin{equation}
\begin{aligned}
T&=T_{\mathrm{ICP}}\circ T_{\mathrm{coarse}}\in SE(3),\\
\hat{\mathbf q}_i&=T(\mathbf q_i),\qquad
\hat Q=\{\hat{\mathbf q}_i\}_{i=1}^{N}.
\end{aligned}
\label{eq:registration}
\end{equation}
Here, $\circ$ denotes transformation composition and $SE(3)$ is the set of 3D rigid transformations. All aligned points are expressed in the reference coordinate system.

\subsection{Local Residual Context \texorpdfstring{(LRC)}{(LRC)}}
\label{sec:residual_context}

Let $\pi(i)$ index the reference point nearest to $\hat{\mathbf q}_i$. The residual $d_i$ is their Euclidean distance divided by $L$.
\begin{equation}
\begin{aligned}
\pi(i)&=\arg\min_{1\leq j\leq M}\|\hat{\mathbf q}_i-\mathbf r_j\|_2,\\
d_i&=\frac{\|\hat{\mathbf q}_i-\mathbf r_{\pi(i)}\|_2}{L},
\end{aligned}
\label{eq:geometric_residual}
\end{equation}
where $L$ is the diagonal length of the reference bounding box and $\|\cdot\|_2$ denotes the Euclidean norm. The nearest reference point $\mathbf r_{\pi(i)}$ serves as the normal anchor for structural comparison.

In Fig.~\ref{fig:motivation}, the normal parallel offset and the anomalous bulge have the same residual at their centers. For the offset, residuals remain nearly constant across neighboring points. For the bulge, they decrease from the tip toward the edges. LRC therefore compares each point's residual with those of its neighbors. We use two neighborhood sizes to extend the receptive field while preserving sensitivity to local changes in residuals.

We spatially sample at most 2,048 support points from $\hat Q$ and compute their residuals using~\eqref{eq:geometric_residual}. For a neighborhood size $k\in\{16,64\}$, let $\mathcal N_k(i)$ be the index set of up to $k$ nearest support points to $\hat{\mathbf q}_i$, excluding the query point itself. Its size is $n_{ik}=|\mathcal N_k(i)|$. For a nonempty neighborhood, the mean residual $\mu_k(i)$ and standard deviation $\sigma_k(i)$ are
\begin{equation}
\begin{aligned}
\mu_k(i)&=\frac{1}{n_{ik}}\sum_{h\in\mathcal N_k(i)}d_h,\\
\sigma_k(i)&=\sqrt{\frac{1}{n_{ik}}\sum_{h\in\mathcal N_k(i)}
\bigl(d_h-\mu_k(i)\bigr)^2+\epsilon},
\end{aligned}
\label{eq:context_statistics}
\end{equation}
where $h$ indexes a support point, $d_h$ is its residual, and $\epsilon=10^{-12}$ provides numerical stability.

The residual context vector $\mathbf c_i$ concatenates the point residual with both sets of neighborhood statistics:
\begin{equation}
\begin{aligned}
\mathbf c_i=\bigl[&d_i,\mu_{16}(i),\sigma_{16}(i),d_i-\mu_{16}(i),\\
&\mu_{64}(i),\sigma_{64}(i),d_i-\mu_{64}(i)\bigr]
\in\mathbb R^7.
\end{aligned}
\label{eq:residual_context}
\end{equation}
The mean provides a local residual baseline, the standard deviation measures the dispersion of neighboring residuals, and the centered residual $d_i-\mu_k(i)$ records the point's departure from that baseline. Normal shape variation or registration error can yield large residuals over a region with small centered residuals. Comparing these patterns at two neighborhood sizes helps the scorer separate shared geometric offsets from localized inconsistencies.

\subsection{Multi-scale Structural Discrepancy Encoding \texorpdfstring{(MSDE)}{(MSDE)}}
\label{sec:structural_comparison}

LRC captures how far aligned points lie from the reference and how these distances vary across a neighborhood. A surface patch can exhibit a large geometric residual while retaining its local shape, as in the parallel offset in Fig.~\ref{fig:motivation}. MSDE therefore compares descriptors of corresponding patches at multiple scales, providing local shape information that complements the distance statistics in LRC. We use the frozen MiniSpinNet encoder $E$~\cite{ao2023buffer} pretrained for registration in BUFFER-X~\cite{lim2026toward}, with its three adaptive patch scales. Its rotation-invariant design supports local shape comparison across poses, a property also used for 3D AD by Reg2Inv~\cite{yu2025registration}. At scale $s\in\{1,2,3\}$, let $\mathcal P_{m,s}^{b}$ denote the $m$-th point patch, with center $\mathbf p_{m,s}^{b}$ and radius $r_s$. The superscript $b\in\{Q,R\}$ identifies the query or reference side. The encoder produces its descriptor $\mathbf v_{m,s}^{b}$ as
\begin{equation}
\mathbf v_{m,s}^{b}=E(\mathcal P_{m,s}^{b})\in\mathbb R^{32}.
\label{eq:patch_descriptor}
\end{equation}

Let $a_s^Q(i)$ and $a_s^R(i)$ denote the query and reference patch indices associated with query point $i$ at scale $s$. We transform query patch centers by $T$ and obtain these indices through spatial nearest-neighbor association:
\begin{equation}
\begin{aligned}
a_s^Q(i)&=\arg\min_m\|\hat{\mathbf q}_i-T(\mathbf p_{m,s}^{Q})\|_2,\\
a_s^R(i)&=\arg\min_m\|\mathbf r_{\pi(i)}-\mathbf p_{m,s}^{R}\|_2.
\end{aligned}
\label{eq:patch_correspondence}
\end{equation}
The selected descriptors are $\mathbf v_{a_s^Q(i),s}^{Q}$ and $\mathbf v_{a_s^R(i),s}^{R}$. Thus, registration determines which normal region is compared with the query, and descriptor comparison measures their structural discrepancy. This correspondence follows spatial alignment rather than descriptor similarity.

At each scale $s$, MSDE applies the same learnable projection $\mathbf W_s\in\mathbb R^{32\times32}$ to the query and reference descriptors. Training this projection with the anomaly objective allows MSDE to compare corresponding patches in a common feature space adapted to anomaly detection.
\begin{equation}
\mathbf u_{i,s}^{b}=\omega_{i,s}^{b}\mathbf W_s\bar{\mathbf v}_{i,s}^{b},
\qquad b\in\{Q,R\}.
\label{eq:descriptor_projection}
\end{equation}
For a valid association, $\omega_{i,s}^{b}=1$ and $\bar{\mathbf v}_{i,s}^{b}=\mathbf v_{a_s^b(i),s}^{b}/\|\mathbf v_{a_s^b(i),s}^{b}\|_2$. For an invalid association, we set $\omega_{i,s}^{b}=0$ and $\bar{\mathbf v}_{i,s}^{b}=\mathbf 0$. The validity criteria are given in Sec.~\ref{sec:msde_validity} of the supplementary material.

Let $\omega_{i,s}=\omega_{i,s}^{Q}\omega_{i,s}^{R}$ indicate that both descriptor associations are valid. The structural comparison feature $\mathbf f_{i,s}$ concatenates the two projected descriptors with two pairwise discrepancy terms:
\begin{equation}
\begin{aligned}
\mathbf f_{i,s}=\bigl[&\mathbf u_{i,s}^{Q},\mathbf u_{i,s}^{R},
 \omega_{i,s}|\mathbf u_{i,s}^{Q}-\mathbf u_{i,s}^{R}|,\\
&\omega_{i,s}\bigl(1-(\bar{\mathbf v}_{i,s}^{Q})^\top
\bar{\mathbf v}_{i,s}^{R}\bigr)\bigr].
\end{aligned}
\label{eq:structural_feature}
\end{equation}
If either association is invalid, both pairwise terms are zero. The scorer retains the residual context and any valid query or reference descriptor.

The two projected descriptors retain the observed and expected local structures. Their elementwise absolute difference compares the learned representations, while cosine dissimilarity measures disagreement in the pretrained descriptor space. Each scale contributes $32+32+32+1=97$ dimensions. Concatenating all three scales gives the structural representation $\mathbf f_i=[\mathbf f_{i,1},\mathbf f_{i,2},\mathbf f_{i,3}]\in\mathbb R^{291}$. Small patches describe fine local changes, while larger patches capture their surrounding shape. Combined with LRC, these features let the scorer assess whether a large residual is accompanied by a structural discrepancy. Figure~\ref{fig:csfn}(b) shows how MSDE constructs and concatenates the per-scale features.

\subsection{Context-Structure Fusion Network \texorpdfstring{(CSFN)}{(CSFN)}}
\label{sec:csfn}

\begin{figure}[!tbp]
    \centering
    \includegraphics[width=\columnwidth,pagebox=cropbox]{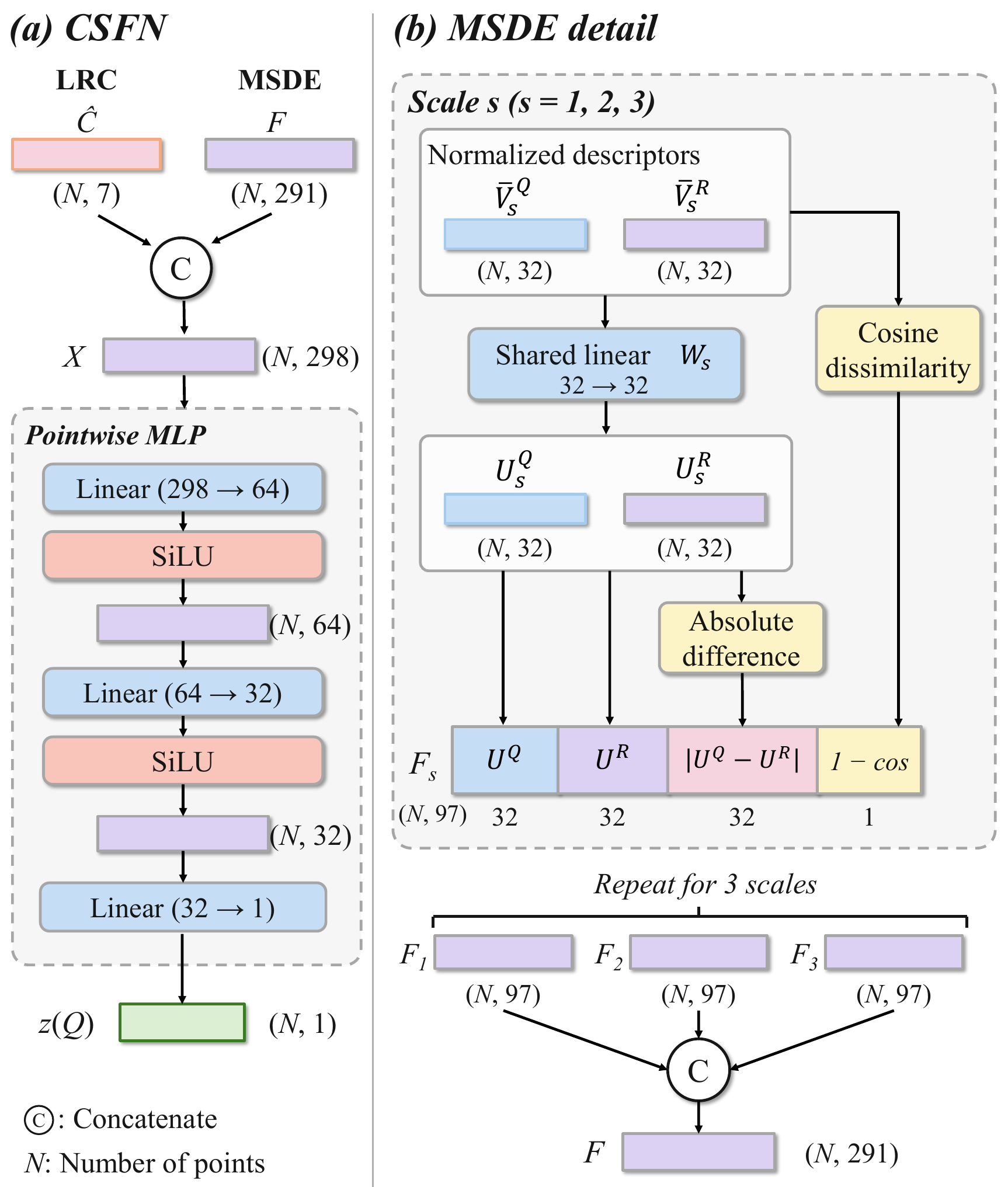}
    \caption{CSFN and MSDE architectures. (a) CSFN fuses LRC and MSDE features for pointwise anomaly scoring. (b) MSDE projects descriptors with $\mathbf W_s$, encodes structural discrepancies at each scale, and concatenates the features across three scales. $N$ denotes the number of query points.}
    \label{fig:csfn}
\end{figure}

CSFN fuses LRC and MSDE features to predict pointwise anomaly scores through a multilayer perceptron (MLP) $F_\theta$ (Fig.~\ref{fig:csfn}(a)), where $\theta$ denotes its trainable weights and biases. We standardize each channel of $\mathbf c_i$ using means and standard deviations estimated from clean training scans, obtaining $\hat{\mathbf c}_i$. These statistics remain fixed at inference, and entries with unavailable neighborhood statistics are set to zero after standardization. The concatenated input vector $\mathbf x_i$ and predicted anomaly logit $z_i$ are
\begin{equation}
\mathbf x_i=[\hat{\mathbf c}_i,\mathbf f_i]\in\mathbb R^{298},
\qquad z_i=F_\theta(\mathbf x_i).
\label{eq:csfn_score}
\end{equation}
The MLP has layer widths $298\rightarrow64\rightarrow32\rightarrow1$, with sigmoid linear unit (SiLU) activations after the two hidden layers. Joint access to both representations lets the scorer interpret a point's residual using both its neighborhood residuals and its structural discrepancy from the normal reference.

The logits form the pointwise anomaly map $\mathbf z(Q)=[z_1,\ldots,z_N]^\top$. For object-level detection, let $K$ be the pooling size and $K'=\min(K,N)$ the number of points retained. If $\mathcal T_{K'}(Q)$ denotes the index set of the $K'$ highest-scoring points, their average gives
\begin{equation}
S_Q=\frac{1}{K'}\sum_{i\in\mathcal T_{K'}(Q)}z_i.
\label{eq:object_score}
\end{equation}

\subsection{Training Objective}
\label{sec:training_inference}

From a normal training sample in $\mathcal D$, we form a pair consisting of an augmented clean sample $Q^{\mathrm{clean}}$ and its synthetic anomalous counterpart $Q^{\mathrm{defect}}$. The anomaly is generated by applying a smooth, signed displacement along locally consistent surface normals within a connected region. The pair shares Gaussian noise and rigid pose augmentation. Figure~\ref{fig:synthetic_pairs} in the supplementary material shows examples of these training pairs. The anomaly label vector $\mathbf y=[y_1,\ldots,y_N]^\top$ assigns one to modified points and zero to unchanged points; the clean label vector is $\mathbf 0\in\mathbb R^N$. Both scans are independently registered to the same normal reference and processed by the shared feature and scoring modules.

Let $\mathcal I_+$ and $\mathcal I_-$ be the index sets of points sampled for supervision with $y_i=1$ and $y_i=0$ in $Q^{\mathrm{defect}}$, respectively, and let $\mathcal I_0$ index the sampled points in $Q^{\mathrm{clean}}$. We denote the logits from the anomalous and clean scans by $z_i^{\mathrm{defect}}$ and $z_i^{\mathrm{clean}}$, respectively. With the predicted probabilities $\hat y_i^{\mathrm{defect}}=\mathrm{sigmoid}(z_i^{\mathrm{defect}})$ and $\hat y_i^{\mathrm{clean}}=\mathrm{sigmoid}(z_i^{\mathrm{clean}})$, we define the positive, negative, and clean binary cross-entropy (BCE) losses as
{
\begin{equation}
\begin{aligned}
\mathcal L_{\mathrm{pos}}&=-\frac{1}{|\mathcal I_+|}
\sum_{i\in\mathcal I_+}\log\hat y_i^{\mathrm{defect}},\\
\mathcal L_{\mathrm{neg}}&=-\frac{1}{|\mathcal I_-|}
\sum_{i\in\mathcal I_-}\log\bigl(1-\hat y_i^{\mathrm{defect}}\bigr),\\
\mathcal L_{\mathrm{clean}}&=-\frac{1}{|\mathcal I_0|}
\sum_{i\in\mathcal I_0}\log\bigl(1-\hat y_i^{\mathrm{clean}}\bigr).
\end{aligned}
\label{eq:training_losses}
\end{equation}
}
The total objective is
{
\begin{equation}
\mathcal L=\lambda_1\mathcal L_{\mathrm{pos}}
+\lambda_2\mathcal L_{\mathrm{neg}}
+\lambda_3\mathcal L_{\mathrm{clean}},
\label{eq:total_loss}
\end{equation}
}
where $\lambda_1$, $\lambda_2$, and $\lambda_3$ weight the positive, negative, and clean losses, respectively.
The positive term promotes high scores at synthetic anomalies, while the two normal terms suppress scores in unchanged regions and clean scans. We jointly optimize the CSFN parameters $\theta$ and the MSDE projections $\{\mathbf W_s\}_{s=1}^{3}$, keeping the registration module and descriptor encoder frozen.

%% file: sec/4_expriments.tex
\begin{table*}[t]
    \centering
    \caption{Quantitative comparison of mean object-level and point-level AUROC (\%) on Real3D-AD and Anomaly-ShapeNet. Bold and underlined values denote the best and second-best results, respectively. See the supplement (Sec. \ref{sec:per_category_results}) for per-category scores.}
    \label{tab:main_comparison}
    \small
    \setlength{\tabcolsep}{4pt}
    \renewcommand{\arraystretch}{1.03}
    \begin{tabular*}{\textwidth}{@{\extracolsep{\fill}}llcrrrr@{}}
        \toprule
        \multirow{2}{*}{Setting} & \multirow{2}{*}{Method} & \multirow{2}{*}{Venue} & \multicolumn{2}{c}{Real3D-AD} & \multicolumn{2}{c}{Anomaly-ShapeNet} \\ 
        \cmidrule(lr){4-5}\cmidrule(lr){6-7}
        & & & O-AUROC & P-AUROC & O-AUROC & P-AUROC \\
        \midrule
        \multirow{16}{*}{\shortstack[l]{Category-\\specific}} & BTF (Raw)~\cite{horwitz2023back} & CVPRW 2023 & 63.5 & 57.1 & 49.3 & 55.0 \\
         & BTF (FPFH)~\cite{horwitz2023back} & CVPRW 2023 & 60.3 & 73.3 & 52.8 & 62.8 \\
         & M3DM~\cite{wang2023multimodal} & CVPR 2023 & 55.2 & 63.1 & 55.2 & 61.6 \\
         & PatchCore (FPFH)~\cite{roth2022towards} & CVPR 2022 & 59.3 & 68.2 & 56.8 & 58.0 \\
         & PatchCore (PointMAE)~\cite{roth2022towards} & CVPR 2022 & 59.5 & 62.0 & 56.2 & 57.7 \\
         & CPMF~\cite{cao2024complementary} & PR 2024 & 58.6 & 75.9 & 55.9 & 57.3 \\
         & IMRNet~\cite{li2024towards} & CVPR 2024 & 72.5 & -- & 65.9 & 65.0 \\
         & Reg3D-AD~\cite{liu2023real3d} & NeurIPS 2023 & 70.4 & 70.5 & 57.2 & 66.8 \\
         & Group3AD~\cite{zhu2024towards} & ACM MM 2024 & 75.1 & 73.5 & -- & -- \\
         & R3D-AD~\cite{zhou2024r3d} & ECCV 2024 & 73.4 & 59.2 & 74.9 & -- \\
         & ISMP~\cite{liang2025look} & AAAI 2025 & 75.7 & 83.6 & -- & 69.1 \\
         & PO3AD~\cite{ye2025po3ad} & CVPR 2025 & 76.6 & 65.0 & 83.9 & 89.9 \\
         & PASDF~\cite{zheng2025bridging} & ICCV 2025 & 80.2 & 74.5 & 90.7 & 89.6 \\
         & Reg2Inv~\cite{yu2025registration} & NeurIPS 2025 & 78.1 & 87.9 & 86.1 & 88.2 \\
         & Simple3D~\cite{cheng2026highresolution} & AAAI 2026 & 80.4 & \underline{92.3} & 86.0 & \underline{92.9} \\
         & CASL~\cite{zha2026casl} & AAAI 2026 & \underline{82.3} & 88.2 & 88.7 & 89.9 \\
        \midrule
        \multirow{3}{*}{Unified} & MC3D-AD~\cite{cheng2025mc3d} & IJCAI 2025 & 78.2 & 76.8 & 84.2 & 75.9 \\
         & SeDiR~\cite{kim2026semantically} & CVPR 2026 & 81.0 & 80.6 & \underline{93.3} & 81.1 \\
         & \textbf{\ours \ (Ours)} & -- & \textbf{92.0} & \textbf{95.6} & \textbf{95.7} & \textbf{98.2} \\
        \bottomrule
    \end{tabular*}
\end{table*}

\section{Experiments}
\label{sec:experiments}

\begin{figure*}[!t]
    \centering
    \includegraphics[width=0.95\textwidth]{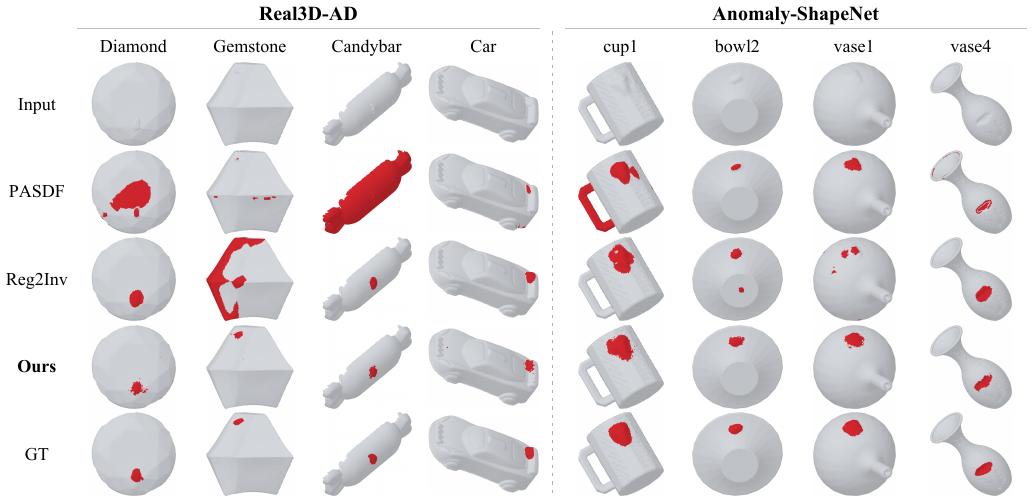}
    \caption{Qualitative comparison of PASDF, Reg2Inv, and \ours{} on Real3D-AD (left) and Anomaly-ShapeNet (right). From top to bottom, the rows show the input point clouds, PASDF predictions, Reg2Inv predictions, \ours{} predictions, and ground truth. Red indicates anomalous points and gray indicates normal points. Each prediction is binarized at its per-scan F1-max threshold.}
    \label{fig:qualitative}
\end{figure*}

\subsection{Experimental Setup}
\label{sec:experimental_setup}

\textbf{Datasets.} We evaluate \ours \ on Real3D-AD~\cite{liu2023real3d} and Anomaly-ShapeNet~\cite{li2024towards}, which contain real scans and synthetic 3D shapes, respectively. Our evaluation covers 12 categories and 1,206 test scans from Real3D-AD, and 40 categories and 1,312 test scans from Anomaly-ShapeNet. For each dataset, we train one model jointly across all categories using normal training samples and their synthetic anomalous counterparts.

\textbf{Evaluation metrics.} Following the benchmark protocols~\cite{liu2023real3d,li2024towards}, we report the area under the receiver operating characteristic curve (AUROC) at the object level (O-AUROC) for detection and at the point level (P-AUROC) for localization. We compute each metric per category and then average equally across categories, reporting the two datasets separately.

\textbf{Implementation details.} We freeze the pretrained BUFFER-X~\cite{lim2026toward} registration model and its descriptor encoder, and jointly train the MSDE projections and CSFN with AdamW for 5,000 updates. We train on a single NVIDIA GeForce RTX~4090 GPU. The learning rate is $10^{-3}$ and weight decay is $10^{-4}$. After registration, we randomly sample 4,096 points for pointwise supervision. The loss weights in~\eqref{eq:total_loss} are $\lambda_1=0.25$, $\lambda_2=0.25$, and $\lambda_3=0.50$.
We follow PASDF's normal-reference selection and coordinate preprocessing protocols~\cite{zheng2025bridging}. LRC uses at most 2,048 support points and neighborhoods of 16 and 64 points. Object-level scores use $K=1000$ for Real3D-AD and $K=1$ for Anomaly-ShapeNet in~\eqref{eq:object_score}. Anomaly synthesis is detailed in the supplementary material.

\subsection{Comparison with Existing Methods}
\label{sec:main_results}

We compare \ours \ with category-specific and unified 3D AD methods. The category-specific baselines include BTF~\cite{horwitz2023back}, M3DM~\cite{wang2023multimodal}, PatchCore~\cite{roth2022towards}, CPMF~\cite{cao2024complementary}, IMRNet~\cite{li2024towards}, Group3AD~\cite{zhu2024towards}, R3D-AD~\cite{zhou2024r3d}, ISMP~\cite{liang2025look}, PO3AD~\cite{ye2025po3ad}, Reg3D-AD~\cite{liu2023real3d}, PASDF~\cite{zheng2025bridging}, Reg2Inv~\cite{yu2025registration}, Simple3D~\cite{cheng2026highresolution}, and CASL~\cite{zha2026casl}. We include the Raw and fast point feature histogram (FPFH) \cite{rusu2009fast} variants of BTF and the FPFH and PointMAE variants of PatchCore. MC3D-AD~\cite{cheng2025mc3d} and SeDiR~\cite{kim2026semantically} serve as unified baselines that share one model across categories, as does \ours. Comparisons with Reg3D-AD, PASDF, and Reg2Inv examine the benefits of our learned scoring framework relative to existing registration-based approaches. Table~\ref{tab:main_comparison} reports mean O-AUROC and P-AUROC across categories for each method.

On Real3D-AD, \ours \ achieves mean O-AUROC and P-AUROC of 92.0\% and 95.6\%, respectively. The next-highest scores in Table~\ref{tab:main_comparison} are 82.3\% O-AUROC for CASL~\cite{zha2026casl} and 92.3\% P-AUROC for Simple3D~\cite{cheng2026highresolution}. Compared with the registration-based methods Reg3D-AD, PASDF, and Reg2Inv, \ours \ improves O-AUROC by at least 11.8 percentage points and P-AUROC by at least 7.7 points.

On Anomaly-ShapeNet, \ours \ achieves mean O-AUROC and P-AUROC of 95.7\% and 98.2\%, respectively, across 40 categories. These scores exceed the strongest baselines in Table~\ref{tab:main_comparison} by 2.4 points in O-AUROC (SeDiR) and 5.3 points in P-AUROC (Simple3D). \ours \ also outperforms all three registration-based baselines on both metrics, exceeding PASDF by 5.0 and 8.6 points in O-AUROC and P-AUROC, respectively. These gains are consistent with our key insight, as verified in Sec.~\ref{sec:parameter_analysis} under identical registration results, particularly for pointwise localization.

Figure~\ref{fig:qualitative} qualitatively compares PASDF, Reg2Inv, and \ours{} on both datasets. On Real3D-AD, PASDF produces extensive false positives on Candybar and Diamond, while Reg2Inv produces spurious responses along several edges of Gemstone. \ours{} reduces these false positives and detects more of the annotated anomalous region on Car than PASDF. On Anomaly-ShapeNet, PASDF incorrectly flags the handle of cup1, while Reg2Inv flags a normal region of bowl2 and detects only part of the annotated anomaly on vase1. \ours{} leaves the cup handle largely unmarked and detects the anomalous region on vase1 more completely. \ours{} yields fewer false positives in normal regions than other registration-based methods, supporting our hypothesis that residual magnitude alone is insufficient for anomaly scoring.

\begin{table*}[t]
    \begin{minipage}[t]{\columnwidth}
    \centering
    \caption{LRC and MSDE ablations. Values are O/P-AUROC (\%); O and P denote object-level and point-level evaluation. All configurations retain the center residual $d_i$ and use CSFN for learned scoring.}
    \label{tab:modules}
    \footnotesize
    \setlength{\tabcolsep}{4pt}
    \begin{tabular*}{\columnwidth}{@{\extracolsep{\fill}}cccccc@{}}
        \toprule
        \multirow{2}{*}[-0.6ex]{LRC} & \multirow{2}{*}[-0.6ex]{MSDE} & \multicolumn{2}{c}{Real3D-AD} & \multicolumn{2}{c}{Anomaly-ShapeNet} \\
        \cmidrule(lr){3-4}\cmidrule(lr){5-6}
        & & \makebox[3.7em]{O} & \makebox[3.7em]{P} & \makebox[3.7em]{O} & \makebox[3.7em]{P} \\
        \midrule
        & & 73.85 & 78.19 & 95.68 & 89.25 \\
        $\checkmark$ & & 82.38 & 92.30 & 92.27 & 90.49 \\
        & $\checkmark$ & 88.73 & 88.91 & 95.52 & 96.47 \\
        $\checkmark$ & $\checkmark$ & \textbf{92.01} & \textbf{95.63} & \textbf{95.73} & \textbf{98.18} \\
        \bottomrule
    \end{tabular*}
    \end{minipage}\hfill
    \begin{minipage}[t]{\columnwidth}
    \centering
    \caption{Descriptor choice with LRC fixed (O/P-AUROC, \%). MiniSpinNet is the original configuration.}
    \label{tab:descriptor_encoder}
    \footnotesize
    \setlength{\tabcolsep}{2.2pt}
    \begin{tabular*}{\columnwidth}{@{\extracolsep{\fill}}lcccc@{}}
        \toprule
        \multicolumn{1}{@{}c}{\multirow{2}{*}[-0.6ex]{Descriptor}} & \multicolumn{2}{c}{Real3D-AD} & \multicolumn{2}{c}{Anomaly-ShapeNet} \\
        \cmidrule(lr){2-3}\cmidrule(lr){4-5}
        & \makebox[3.2em]{O} & \makebox[3.2em]{P} & \makebox[3.2em]{O} & \makebox[3.2em]{P} \\
        \midrule
        PointMAE (three-scale pooling) & 84.71 & 93.07 & 95.36 & 98.20 \\
        PointMAE (pointwise) & 84.74 & 93.14 & 95.44 & \textbf{98.21} \\
        MiniSpinNet (three scales) & \textbf{92.01} & \textbf{95.63} & \textbf{95.73} & 98.18 \\
        \bottomrule
    \end{tabular*}
    \end{minipage}
\end{table*}

\subsection{Ablation Studies}
\label{sec:ablation}

\textbf{Module contributions.} Table~\ref{tab:modules} compares four input configurations for CSFN. With both LRC and MSDE removed, CSFN predicts anomaly scores from the center residual $d_i$ alone. This baseline yields the lowest O-AUROC and P-AUROC on Real3D-AD and the lowest P-AUROC on Anomaly-ShapeNet. Adding either LRC or MSDE improves both metrics on Real3D-AD and P-AUROC on Anomaly-ShapeNet. Combining LRC and MSDE gives the highest O-AUROC and P-AUROC on both datasets. These results demonstrate the importance of both LRC and MSDE and their complementary roles in anomaly scoring.

\textbf{Choice of local descriptor.}
With registration and LRC held fixed, we compare MiniSpinNet~\cite{ao2023buffer,lim2026toward} with the frozen PointMAE encoder used in M3DM~\cite{wang2023multimodal}. For PointMAE, we evaluate two variants: direct comparison of features at corresponding points and comparison after spatial pooling at the same three scales used by MiniSpinNet. All three configurations achieve similar performance on Anomaly-ShapeNet (Table~\ref{tab:descriptor_encoder}). On Real3D-AD, MiniSpinNet outperforms the stronger of the two PointMAE variants by 7.27 and 2.49 points in O-AUROC and P-AUROC, respectively. This advantage on real scans may be partly explained by the rotation invariance of MiniSpinNet descriptors~\cite{liang2026lightweight}.

\subsection{Analysis}
\label{sec:parameter_analysis}

\textbf{Direct distance scoring.} In the residual-only ablation (Table~\ref{tab:modules}), the geometric residual $d_i$ is the input to a trained CSFN. Here $d_i$ itself serves as the pointwise anomaly score. Table~\ref{tab:fixed_learned} compares this distance, descriptor cosine dissimilarity averaged over valid scales, and the learned scores from \ours{} on the same aligned points. Relative to geometric distance, \ours \ improves P-AUROC by 12.56 points on Real3D-AD and 8.93 points on Anomaly-ShapeNet. The corresponding O-AUROC gains are 9.03 and 0.05 points. These results support learning anomaly scores from contextual and structural evidence, with the largest gains in pointwise localization.

\begin{figure}[!tbp]
    \centering
    \includegraphics[width=\columnwidth]{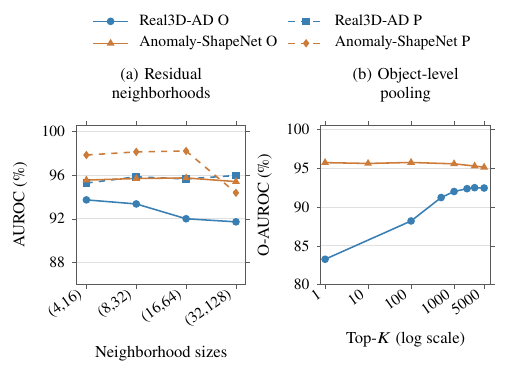}
    \caption{Parameter analysis of (a) residual neighborhood sizes and (b) object-level Top-$K$ pooling (logarithmic horizontal axis). O/P denote object-level and point-level AUROC. Default neighborhoods are $(16,64)$; default $K$ values are 1000 and 1 on Real3D-AD and Anomaly-ShapeNet, respectively.}
    \label{fig:parameter_analysis}
\end{figure}

\textbf{Residual neighborhood size.} Figure~\ref{fig:parameter_analysis}(a) compares four residual neighborhood size pairs: $(4,16)$, $(8,32)$, $(16,64)$, and $(32,128)$. On Real3D-AD, O-AUROC decreases from 93.73\% at $(4,16)$ to 91.72\% at $(32,128)$, whereas P-AUROC ranges from 95.26\% to 95.95\% and peaks at $(32,128)$. On Anomaly-ShapeNet, $(16,64)$ yields the highest O-AUROC and P-AUROC, at 95.73\% and 98.18\%, respectively, compared with 95.53\% and 97.82\% at $(4,16)$. Increasing the neighborhood sizes to $(32,128)$ reduces P-AUROC to 94.37\%. Overall, performance is relatively stable up to $(16,64)$, while the largest neighborhoods offer no consistent improvement across datasets and metrics. These declines may arise because larger neighborhoods mix distinct local structures and dilute localized residual cues. We retain the default $(16,64)$.

\textbf{\boldmath Top-$K$ pooling.} We vary $K$ with fixed pointwise anomaly scores (Fig.~\ref{fig:parameter_analysis}(b)). On Real3D-AD, O-AUROC increases from 83.27\% at $K=1$ to 92.01\% at $K=1000$. It then levels off, reaching 92.49\% at $K=3000$ and 92.45\% at $K=5000$. On Anomaly-ShapeNet, O-AUROC is similar at $K=1$ (95.73\%) and $K=100$ (95.75\%), but decreases to 95.29\% at $K=3000$ and 95.13\% at $K=5000$. Object-level detection performance is therefore more sensitive to $K$ on Real3D-AD. One possible explanation is that real scans are noisier than synthetic shapes and thus more likely to produce isolated high scores. Averaging over more points can reduce their influence. We retain the default values of $1000$ for Real3D-AD and $1$ for Anomaly-ShapeNet.

\begin{table}[!t]
    \centering
    \caption{Fixed distances and learned scoring. Values are AUROC (\%); O and P denote object-level and point-level evaluation.}
    \label{tab:fixed_learned}
    \footnotesize
    \setlength{\tabcolsep}{4pt}
    \begin{tabular*}{\columnwidth}{@{\extracolsep{\fill}}lcccc@{}}
        \toprule
        \multicolumn{1}{@{}c}{\multirow{2}{*}[-0.6ex]{Scoring rule}} & \multicolumn{2}{c}{Real3D-AD} & \multicolumn{2}{c}{Anomaly-ShapeNet} \\
        \cmidrule(lr){2-3}\cmidrule(lr){4-5}
        & \makebox[3.7em]{O} & \makebox[3.7em]{P} & \makebox[3.7em]{O} & \makebox[3.7em]{P} \\
        \midrule
        Geometric distance & 82.98 & 83.07 & 95.68 & 89.25 \\
        Descriptor distance & 59.07 & 82.19 & 68.35 & 90.55 \\
        \ours \ & \textbf{92.01} & \textbf{95.63} & \textbf{95.73} & \textbf{98.18} \\
        \bottomrule
    \end{tabular*}
\end{table}

\textbf{Computational efficiency.}
The median inference time of \ours{} on Real3D-AD is 3.1\,s per scan, comparable to that of other registration-based methods~\cite{yu2025registration,zheng2025bridging}. However, inference cost remains a common limitation of these approaches.

%% file: sec/5_conlusion.tex
\section{Conclusion}
\label{sec:conclusion}

We presented \ours, a unified framework that learns to interpret registration discrepancies for multi-category 3D anomaly detection. By combining local residual context with multi-scale structural discrepancies, \ours \ learns pointwise anomaly scores using one shared model per dataset. Experiments on Real3D-AD and Anomaly-ShapeNet show improvements over the compared registration-based methods in both detection and localization. On both datasets, \ours\ achieves the highest mean O-AUROC and P-AUROC. Ablation studies demonstrate the importance and complementary roles of local residual context and multi-scale structural discrepancies. These findings support learning anomaly scores from contextual and structural evidence after registration.

%% file: sec/6_llm_usage.tex
\section*{LLM Usage Statement}
ChatGPT was used to assist with language editing, related work and experimental planning and analysis. The authors take full responsibility for all submitted content.

%% file: sec/X_suppl.tex
\clearpage
\onecolumn
\makeatletter
\setlength{\@fptop}{0pt}
\makeatother
\setcounter{page}{1}
\renewcommand{\thepage}{S\arabic{page}}
\appendix
\setcounter{table}{0}
\renewcommand{\thetable}{A\arabic{table}}
\setcounter{figure}{0}
\renewcommand{\thefigure}{A\arabic{figure}}
\renewcommand{\theHfigure}{supp.\arabic{figure}}
% Use the CVPR supplementary title styling in a full-width table appendix.
\begingroup
\centering
\Large
\textbf{\thetitle}\\
\vspace{0.5em}Supplementary Material\par
\vspace{1.0em}
\endgroup

\section{\texorpdfstring{MSDE Descriptor Validity}{MSDE Descriptor Validity}}
\label{sec:msde_validity}

We check each query and reference association separately at each scale $s$. The selected descriptor must be finite and have Euclidean norm greater than $10^{-8}$. We measure patch-center distances in the reference coordinate system. On the query side, the transformed patch center must lie within $r_s/2$ of $\hat{\mathbf q}_i$. On the reference side, the patch center must lie within $r_s/2$ of $\mathbf r_{\pi(i)}$, and the reference anchor must satisfy $\|\hat{\mathbf q}_i-\mathbf r_{\pi(i)}\|_2\leq r_s$. At each scale, stacking the pointwise descriptors for side $b\in\{Q,R\}$ gives $\bar{\mathbf V}_s^b\in\mathbb R^{N\times32}$, with one row per query point and zero rows for invalid associations. The projection matrix $\mathbf W_s\in\mathbb R^{32\times32}$ is shared by the two sides.

\section{Additional Experimental Details}
\label{sec:implementation_details}

\textbf{Anomaly synthesis.} Both scans of each training pair share Gaussian noise and a random rotation and translation. The radius and magnitude of the anomalous displacement are sampled from 3--8\% and 0.2--1\% of the source bounding-box diagonal, respectively.

\begin{figure}[!htbp]
    \centering
    \includegraphics[width=\linewidth]{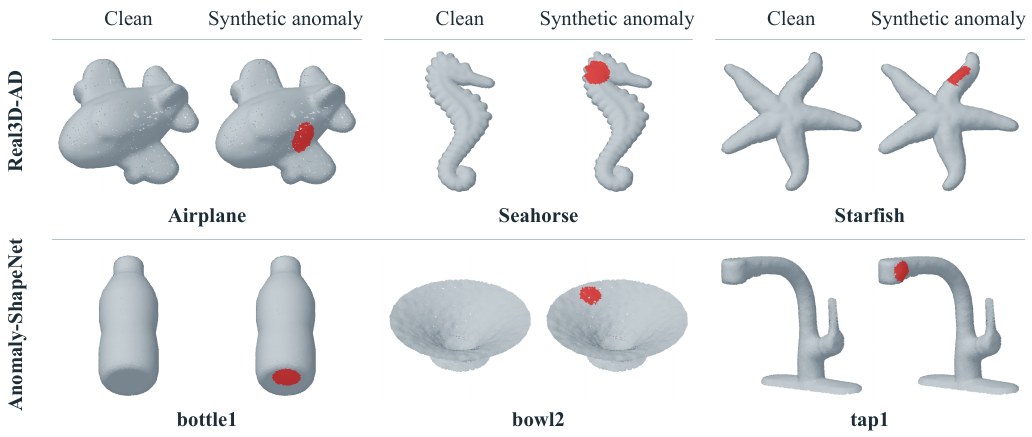}
    \caption{Synthetic training pairs from Real3D-AD (top) and Anomaly-ShapeNet (bottom). Each clean point cloud is shown alongside its synthetic anomalous counterpart from the same viewpoint before registration. Red marks the points displaced during synthesis.}
    \label{fig:synthetic_pairs}
\end{figure}

% \clearpage
\section{Per-Category Results}
\label{sec:per_category_results}

Tables~\ref{tab:real3d_object}--\ref{tab:shapenet_point} give the available per-category scores for Table~\ref{tab:main_comparison}. Means are unweighted averages over categories. For BTF, M3DM, PatchCore, CPMF, and Reg3D-AD, we use the benchmark results of Li et al.~\cite{li2024towards} for object-level AUROC and for point-level AUROC on Anomaly-ShapeNet. Their Real3D-AD point-level scores follow Liang et al.~\cite{liang2025look}, except for CPMF, which follows Zhu et al.~\cite{zhu2024towards}. The Anomaly-ShapeNet point-level results for MC3D-AD are from Kim et al.~\cite{kim2026semantically}.

\begin{table*}[!htp]
    \centering
    \caption{Per-category object-level AUROC (\%) on Real3D-AD. Bold and underlined values denote the best and second-best results, respectively.}
    \label{tab:real3d_object}
    \fontsize{8}{9.0}\selectfont
    \setlength{\tabcolsep}{2.0pt}
    \renewcommand{\arraystretch}{1.0}
    \begin{tabular*}{\textwidth}{@{\extracolsep{\fill}}lrrrrrrrrrrrrr@{}}
        \toprule
        Method & Airplane & Car & Candybar & Chicken & Diamond & Duck & Fish & Gemstone & Seahorse & Shell & Starfish & Toffees & Mean \\
        \midrule
        \multicolumn{14}{@{}l}{\textit{Category-specific}} \\
        BTF(Raw) & 73.0 & 64.7 & 53.9 & 78.9 & 70.7 & 69.1 & 60.2 & 68.6 & 59.6 & 39.6 & 53.0 & 70.3 & 63.5 \\
        BTF(FPFH) & 52.0 & 56.0 & 63.0 & 43.2 & 54.5 & 78.4 & 54.9 & 64.8 & 77.9 & 75.4 & 57.5 & 46.2 & 60.3 \\
        M3DM & 43.4 & 54.1 & 55.2 & 68.3 & 60.2 & 43.3 & 54.0 & 64.4 & 49.5 & 69.4 & 55.1 & 45.0 & 55.2 \\
        PatchCore(FPFH) & \textbf{88.2} & 59.0 & 54.1 & 83.7 & 57.4 & 54.6 & 67.5 & 37.0 & 50.5 & 58.9 & 44.1 & 56.5 & 59.3 \\
        PatchCore(PointMAE) & 72.6 & 49.8 & 66.3 & 82.7 & 78.3 & 48.9 & 63.0 & 37.4 & 53.9 & 50.1 & 51.9 & 58.5 & 59.5 \\
        CPMF & 70.1 & 55.1 & 55.2 & 50.4 & 52.3 & 58.2 & 55.8 & 58.9 & 72.9 & 65.3 & 70.0 & 39.0 & 58.6 \\
        IMRNet & 76.2 & 71.1 & 75.5 & 78.0 & 90.5 & 51.7 & 88.0 & 67.4 & 60.4 & 66.5 & 67.4 & 77.4 & 72.5 \\
        Reg3D-AD & 71.6 & 69.7 & 68.5 & \underline{85.2} & 90.0 & 58.4 & 91.5 & 41.7 & 76.2 & 58.3 & 50.6 & 82.7 & 70.4 \\
        Group3AD & 74.4 & 72.8 & 84.7 & 78.6 & 93.2 & 67.9 & \underline{97.6} & 53.9 & 84.1 & 58.5 & 56.2 & 79.6 & 75.1 \\
        R3D-AD & 77.2 & 71.3 & 69.6 & 71.4 & 68.5 & \textbf{90.9} & 69.2 & 66.5 & 72.0 & 84.0 & 70.1 & 70.3 & 73.4 \\
        ISMP & 85.8 & 73.1 & 85.2 & 71.4 & 94.8 & 71.2 & 94.5 & 46.8 & 72.9 & 62.3 & 66.0 & 84.2 & 75.7 \\
        PO3AD & 80.4 & 65.4 & 78.5 & 68.6 & 80.1 & 82.0 & 85.9 & 69.3 & 75.6 & 80.0 & 75.8 & 77.1 & 76.6 \\
        PASDF & 62.8 & \underline{95.9} & 78.8 & 73.9 & 89.4 & 65.8 & \textbf{98.9} & 63.4 & \textbf{100.0} & \underline{85.0} & 61.7 & 86.6 & 80.2 \\
        Reg2Inv & 81.8 & 75.8 & \textbf{100.0} & \textbf{94.4} & \textbf{100.0} & 75.0 & 67.2 & 73.5 & 53.2 & 69.2 & 84.1 & 62.6 & 78.1 \\
        Simple3D & 76.5 & \textbf{98.1} & 65.1 & 82.6 & \textbf{100.0} & 77.8 & 91.2 & 70.4 & 93.0 & 51.4 & 69.6 & 88.8 & 80.4 \\
        CASL & 80.8 & 79.9 & 84.8 & 65.7 & 97.6 & 83.6 & 93.5 & \underline{76.9} & 64.3 & 79.1 & \underline{89.3} & \underline{92.4} & \underline{82.3} \\
        \midrule
        \multicolumn{14}{@{}l}{\textit{Unified}} \\
        MC3D-AD & 85.0 & 74.9 & 83.0 & 71.5 & 95.5 & 83.1 & 86.5 & 56.0 & 71.6 & 80.3 & 76.6 & 73.8 & 78.2 \\
        SeDiR & \underline{86.0} & 78.3 & 81.9 & 72.9 & 94.8 & 86.2 & 93.8 & 62.7 & 67.4 & 77.9 & 85.4 & 84.5 & 81.0 \\
        \textbf{\ours \ (Ours)} & 83.3 & 85.6 & \underline{99.5} & 83.5 & \underline{98.8} & \underline{89.8} & 93.7 & \textbf{85.0} & \underline{96.2} & \textbf{93.9} & \textbf{96.5} & \textbf{98.3} & \textbf{92.0} \\
        \bottomrule
    \end{tabular*}
\end{table*}

\begin{table*}[!htp]
    \centering
    \caption{Per-category point-level AUROC (\%) on Real3D-AD. For R3D-AD, Zhou et al.~\cite{zhou2024r3d} report only the overall P-AUROC of the full model. Bold and underlined values denote the best and second-best results, respectively.}
    \label{tab:real3d_point}
    \fontsize{8}{9.0}\selectfont
    \setlength{\tabcolsep}{2.0pt}
    \renewcommand{\arraystretch}{1.0}
    \begin{tabular*}{\textwidth}{@{\extracolsep{\fill}}lrrrrrrrrrrrrr@{}}
        \toprule
        Method & Airplane & Car & Candybar & Chicken & Diamond & Duck & Fish & Gemstone & Seahorse & Shell & Starfish & Toffees & Mean \\
        \midrule
        \multicolumn{14}{@{}l}{\textit{Category-specific}} \\
        BTF(Raw) & 56.4 & 64.7 & 73.5 & 60.9 & 56.3 & 60.1 & 51.4 & 59.7 & 52.0 & 48.9 & 39.2 & 62.3 & 57.1 \\
        BTF(FPFH) & 73.8 & 70.8 & 86.4 & 73.5 & 88.2 & 87.5 & 70.9 & 89.1 & 51.2 & 57.1 & 50.1 & 81.5 & 73.3 \\
        M3DM & 54.7 & 60.2 & 67.9 & 67.8 & 60.8 & 66.7 & 60.6 & 67.4 & 56.0 & 73.8 & 53.2 & 68.2 & 63.1 \\
        PatchCore(FPFH) & 56.2 & 75.4 & 78.0 & 42.9 & 82.8 & 26.4 & 82.9 & 91.0 & 73.9 & 73.9 & 60.6 & 74.7 & 68.2 \\
        PatchCore(PointMAE) & 56.9 & 60.9 & 62.7 & 72.9 & 71.8 & 52.8 & 71.7 & 44.4 & 63.3 & 70.9 & 58.0 & 58.0 & 62.0 \\
        CPMF & 61.8 & 83.6 & 73.4 & 55.9 & 75.3 & 71.9 & \textbf{98.8} & 44.9 & \underline{96.2} & 72.5 & 80.0 & 95.9 & 75.9 \\
        Reg3D-AD & 63.1 & 71.8 & 72.4 & 67.6 & 83.5 & 50.3 & 82.6 & 54.5 & 81.7 & 81.1 & 61.7 & 75.9 & 70.5 \\
        Group3AD & 63.6 & 74.5 & 73.8 & 75.9 & 86.2 & 63.1 & 83.6 & 56.4 & 82.7 & 79.8 & 62.5 & 80.3 & 73.5 \\
        R3D-AD & -- & -- & -- & -- & -- & -- & -- & -- & -- & -- & -- & -- & 59.2 \\
        ISMP & 75.3 & 83.6 & 90.7 & 79.8 & 92.6 & 87.6 & 88.6 & 85.7 & 81.3 & 83.9 & 64.1 & 89.5 & 83.6 \\
        PO3AD & 71.5 & 57.4 & 76.3 & 56.3 & 61.4 & 64.4 & 87.7 & 53.0 & 58.5 & 54.2 & 65.9 & 72.9 & 65.0 \\
        PASDF & 77.7 & 80.2 & 54.6 & 76.8 & 69.9 & 78.2 & 83.7 & 65.4 & 88.7 & 64.8 & 70.3 & 83.8 & 74.5 \\
        Reg2Inv & \textbf{92.3} & \underline{94.4} & \underline{96.9} & \textbf{91.0} & 97.9 & 93.7 & 84.6 & 90.7 & 64.5 & \underline{90.6} & 84.0 & 73.7 & 87.9 \\
        Simple3D & 88.1 & \textbf{99.2} & 96.2 & 86.1 & \textbf{99.0} & \textbf{96.6} & 96.2 & \underline{97.3} & 94.2 & 71.6 & \underline{85.8} & \underline{96.8} & \underline{92.3} \\
        CASL & 84.2 & 90.5 & 93.2 & 71.3 & \underline{98.8} & 89.5 & 93.5 & 91.6 & 81.4 & 87.3 & 83.9 & 93.7 & 88.2 \\
        \midrule
        \multicolumn{14}{@{}l}{\textit{Unified}} \\
        MC3D-AD & 62.8 & 81.9 & 91.0 & 64.0 & 94.2 & 82.2 & 93.2 & 45.8 & 65.9 & 77.8 & 69.0 & 93.4 & 76.8 \\
        SeDiR & 66.6 & 90.8 & 95.1 & 63.2 & 97.7 & 83.4 & 96.7 & 54.7 & 73.9 & 79.8 & 70.9 & 94.0 & 80.6 \\
        \textbf{\ours \ (Ours)} & \underline{88.4} & 92.5 & \textbf{98.2} & \underline{89.7} & 97.4 & \underline{95.9} & \underline{97.1} & \textbf{98.7} & \textbf{97.1} & \textbf{99.0} & \textbf{96.3} & \textbf{97.4} & \textbf{95.6} \\
        \bottomrule
    \end{tabular*}
\end{table*}

\begin{table*}[!htp]
    \centering
    \caption{Per-category object-level AUROC (\%) on Anomaly-ShapeNet. Bold and underlined values denote the best and second-best results, respectively.}
    \label{tab:shapenet_object}
    \fontsize{8}{9.0}\selectfont
    \setlength{\tabcolsep}{2.0pt}
    \renewcommand{\arraystretch}{0.95}
    \begin{tabular*}{\textwidth}{@{\extracolsep{\fill}}lrrrrrrrrrrrrrr@{}}
        \toprule
        Method & ashtray0 & bag0 & bottle0 & bottle1 & bottle3 & bowl0 & bowl1 & bowl2 & bowl3 & bowl4 & bowl5 & bucket0 & bucket1 & cap0 \\
        \midrule
        \multicolumn{15}{@{}l}{\textit{Category-specific}} \\
        BTF(Raw) & 57.8 & 41.0 & 59.7 & 51.0 & 56.8 & 56.4 & 26.4 & 52.5 & 38.5 & 66.4 & 41.7 & 61.7 & 32.1 & 66.8 \\
        BTF(FPFH) & 42.0 & 54.6 & 34.4 & 54.6 & 32.2 & 50.9 & 66.8 & 51.0 & 49.0 & 60.9 & 69.9 & 40.1 & 63.3 & 61.8 \\
        M3DM & 57.7 & 53.7 & 57.4 & 63.7 & 54.1 & 63.4 & 66.3 & 68.4 & 61.7 & 46.4 & 40.9 & 30.9 & 50.1 & 55.7 \\
        PatchCore(FPFH) & 58.7 & 57.1 & 60.4 & 66.7 & 57.2 & 50.4 & 63.9 & 61.5 & 53.7 & 49.4 & 55.8 & 46.9 & 55.1 & 58.0 \\
        PatchCore(PointMAE) & 59.1 & 60.1 & 51.3 & 60.1 & 65.0 & 52.3 & 62.9 & 45.8 & 57.9 & 50.1 & 59.3 & 59.3 & 56.1 & 58.9 \\
        CPMF & 35.3 & 64.3 & 52.0 & 48.2 & 40.5 & 78.3 & 63.9 & 62.5 & 65.8 & 68.3 & 68.5 & 48.2 & 60.1 & 60.1 \\
        Reg3D-AD & 59.7 & 70.6 & 48.6 & 69.5 & 52.5 & 67.1 & 52.5 & 49.0 & 34.8 & 66.3 & 59.3 & 61.0 & 75.2 & 69.3 \\
        IMRNet & 67.1 & 66.0 & 55.2 & 70.0 & 64.0 & 68.1 & 70.2 & 68.5 & 59.9 & 67.6 & 71.0 & 58.0 & 77.1 & 73.7 \\
        R3D-AD & 83.3 & 72.0 & 73.3 & 73.7 & 78.1 & 81.9 & 77.8 & 74.1 & 76.7 & 74.4 & 65.6 & 68.3 & 75.6 & 82.2 \\
        PO3AD & \textbf{100.0} & 83.3 & 90.0 & 93.3 & 92.6 & 92.2 & 82.9 & 83.3 & 88.1 & 98.1 & 84.9 & 85.3 & 78.7 & 87.7 \\
        PASDF & \textbf{100.0} & \underline{99.5} & \textbf{100.0} & \textbf{100.0} & \textbf{100.0} & \textbf{100.0} & 94.8 & \textbf{100.0} & \textbf{100.0} & 93.3 & 91.2 & 96.8 & 77.5 & 85.2 \\
        Reg2Inv & 90.0 & \textbf{100.0} & \textbf{100.0} & \textbf{100.0} & \textbf{100.0} & \textbf{100.0} & 80.7 & 65.6 & 58.5 & 85.2 & 81.8 & 81.3 & 90.2 & 65.9 \\
        Simple3D & \underline{99.5} & 88.1 & \underline{97.6} & 95.1 & \textbf{100.0} & \textbf{100.0} & 83.0 & 71.1 & 91.1 & 73.0 & 86.3 & 95.9 & 79.0 & 85.2 \\
        CASL & 94.3 & 94.8 & 95.7 & \underline{95.4} & \textbf{100.0} & \textbf{100.0} & 93.3 & \underline{99.3} & \underline{99.6} & 94.4 & 88.8 & \textbf{99.0} & 91.4 & \underline{94.8} \\
        \midrule
        \multicolumn{15}{@{}l}{\textit{Unified}} \\
        MC3D-AD & 96.2 & 80.5 & 79.5 & 70.9 & 75.6 & 93.0 & \textbf{97.8} & 71.9 & 88.5 & 91.1 & 75.4 & 89.8 & 78.4 & 79.3 \\
        SeDiR & 97.6 & 88.6 & 92.9 & 91.6 & \underline{98.4} & \underline{95.9} & \underline{95.6} & 92.6 & 96.7 & \underline{98.5} & \underline{97.9} & 91.1 & \textbf{93.3} & \textbf{98.5} \\
        \textbf{\ours \ (Ours)} & \textbf{100.0} & \textbf{100.0} & \textbf{100.0} & \textbf{100.0} & \textbf{100.0} & \textbf{100.0} & 94.8 & \textbf{100.0} & \textbf{100.0} & \textbf{99.3} & \textbf{100.0} & \underline{97.8} & \underline{92.7} & 84.8 \\
        \bottomrule
    \end{tabular*}
    \par\vspace{7pt}
    \begin{tabular*}{\textwidth}{@{\extracolsep{\fill}}lrrrrrrrrrrrrr@{}}
        \toprule
        Method & cap3 & cap4 & cap5 & cup0 & cup1 & eraser0 & headset0 & headset1 & helmet0 & helmet1 & helmet2 & helmet3 & jar \\
        \midrule
        \multicolumn{14}{@{}l}{\textit{Category-specific}} \\
        BTF(Raw) & 52.7 & 46.8 & 37.3 & 40.3 & 52.1 & 52.5 & 37.8 & 51.5 & 55.3 & 34.9 & 60.2 & 52.6 & 42.0 \\
        BTF(FPFH) & 52.2 & 52.0 & 58.6 & 58.6 & 61.0 & 71.9 & 52.0 & 49.0 & 57.1 & 71.9 & 54.2 & 44.4 & 42.4 \\
        M3DM & 42.3 & 77.7 & 63.9 & 53.9 & 55.6 & 62.7 & 57.7 & 61.7 & 52.6 & 42.7 & 62.3 & 37.4 & 44.1 \\
        PatchCore(FPFH) & 45.3 & 75.7 & 79.0 & 60.0 & 58.6 & 65.7 & 58.3 & 63.7 & 54.6 & 48.4 & 42.5 & 40.4 & 47.2 \\
        PatchCore(PointMAE) & 47.6 & 72.7 & 53.8 & 61.0 & 55.6 & 67.7 & 59.1 & 62.7 & 55.6 & 55.2 & 44.7 & 42.4 & 48.3 \\
        CPMF & 55.1 & 55.3 & 69.7 & 49.7 & 49.9 & 68.9 & 64.3 & 45.8 & 55.5 & 58.9 & 46.2 & 52.0 & 61.0 \\
        Reg3D-AD & 72.5 & 64.3 & 46.7 & 51.0 & 53.8 & 34.3 & 53.7 & 61.0 & 60.0 & 38.1 & 61.4 & 36.7 & 59.2 \\
        IMRNet & 77.5 & 65.2 & 65.2 & 64.3 & 75.7 & 54.8 & 72.0 & 67.6 & 59.7 & 60.0 & 64.1 & 57.3 & 78.0 \\
        R3D-AD & 73.0 & 68.1 & 67.0 & 77.6 & 75.7 & 89.0 & 73.8 & 79.5 & 75.7 & 72.0 & 63.3 & 70.7 & 83.8 \\
        PO3AD & 85.9 & 79.2 & 67.0 & 87.1 & 83.3 & \underline{99.5} & 80.8 & 92.3 & 76.2 & 96.1 & 86.9 & 75.4 & 86.6 \\
        PASDF & 64.9 & 64.6 & 85.3 & 97.1 & 85.7 & 95.2 & \textbf{100.0} & 79.5 & 81.2 & 93.8 & 76.5 & 84.6 & \textbf{100.0} \\
        Reg2Inv & 86.3 & 68.1 & \textbf{90.2} & 73.3 & 93.3 & \textbf{100.0} & \textbf{100.0} & 84.3 & 81.7 & \underline{98.6} & 87.5 & 87.6 & \textbf{100.0} \\
        Simple3D & 86.7 & \underline{91.2} & 80.4 & \textbf{100.0} & 82.4 & \textbf{100.0} & \underline{98.2} & 95.7 & 69.0 & 71.9 & 75.4 & 65.5 & 90.5 \\
        CASL & 82.1 & 77.2 & 55.4 & 99.0 & 64.8 & \underline{99.5} & 83.6 & \underline{96.7} & 78.3 & 89.0 & \underline{88.1} & 86.1 & \underline{99.5} \\
        \midrule
        \multicolumn{14}{@{}l}{\textit{Unified}} \\
        MC3D-AD & 70.1 & 83.5 & 76.1 & 74.3 & \underline{95.2} & 77.6 & 86.2 & 88.6 & 67.2 & \textbf{100.0} & 60.9 & \underline{97.9} & 97.1 \\
        SeDiR & \textbf{98.9} & \textbf{99.6} & \underline{87.4} & \underline{99.5} & \textbf{100.0} & 66.7 & 84.9 & 89.0 & \textbf{96.5} & \textbf{100.0} & 87.0 & \textbf{100.0} & \textbf{100.0} \\
        \textbf{\ours \ (Ours)} & \underline{92.6} & 90.2 & \textbf{90.2} & \textbf{100.0} & \textbf{100.0} & \textbf{100.0} & \textbf{100.0} & \textbf{100.0} & \underline{88.7} & \textbf{100.0} & \textbf{91.0} & 92.4 & \textbf{100.0} \\
        \bottomrule
    \end{tabular*}
    \par\vspace{7pt}
    \begin{tabular*}{\textwidth}{@{\extracolsep{\fill}}lrrrrrrrrrrrrrr@{}}
        \toprule
        Method & phone & shelf0 & tap0 & tap1 & vase0 & vase1 & vase2 & vase3 & vase4 & vase5 & vase7 & vase8 & vase9 & Mean \\
        \midrule
        \multicolumn{15}{@{}l}{\textit{Category-specific}} \\
        BTF(Raw) & 56.3 & 16.4 & 52.5 & 57.3 & 53.1 & 54.9 & 41.0 & 71.7 & 42.5 & 58.5 & 44.8 & 42.4 & 56.4 & 49.3 \\
        BTF(FPFH) & 67.1 & 60.9 & 56.0 & 54.6 & 34.2 & 21.9 & 54.6 & 69.9 & 51.0 & 40.9 & 51.8 & 66.8 & 26.8 & 52.8 \\
        M3DM & 35.7 & 56.4 & 75.4 & 73.9 & 42.3 & 42.7 & 73.7 & 43.9 & 47.6 & 31.7 & 65.7 & 66.3 & 66.3 & 55.2 \\
        PatchCore(FPFH) & 38.8 & 49.4 & 75.3 & 76.6 & 45.5 & 42.3 & 72.1 & 44.9 & 50.6 & 41.7 & 69.3 & 66.2 & 66.0 & 56.8 \\
        PatchCore(PointMAE) & 48.8 & 52.3 & 45.8 & 53.8 & 44.7 & 55.2 & 74.1 & 46.0 & 51.6 & 57.9 & 65.0 & 66.3 & 62.9 & 56.2 \\
        CPMF & 50.9 & 68.5 & 35.9 & 69.7 & 45.1 & 34.5 & 58.2 & 58.2 & 51.4 & 61.8 & 39.7 & 52.9 & 60.9 & 55.9 \\
        Reg3D-AD & 41.4 & 68.8 & 67.6 & 64.1 & 53.3 & 70.2 & 60.5 & 65.0 & 50.0 & 52.0 & 46.2 & 62.0 & 59.4 & 57.2 \\
        IMRNet & 75.5 & 60.3 & 67.6 & 69.6 & 53.3 & 75.7 & 61.4 & 70.0 & 52.4 & 67.6 & 63.5 & 63.0 & 59.4 & 65.9 \\
        R3D-AD & 76.2 & 69.6 & 73.6 & 90.0 & 78.8 & 72.9 & 75.2 & 74.2 & 63.0 & 75.7 & 77.1 & 72.1 & 71.8 & 74.9 \\
        PO3AD & 77.6 & 57.3 & 74.5 & 68.1 & 85.8 & 74.2 & 95.2 & 82.1 & 67.5 & 85.2 & \underline{96.6} & 73.9 & 83.0 & 83.9 \\
        PASDF & \textbf{100.0} & 71.3 & 88.2 & 79.3 & \textbf{100.0} & 92.9 & \textbf{100.0} & 80.6 & 91.2 & \textbf{100.0} & \textbf{100.0} & \underline{92.4} & 83.6 & 90.7 \\
        Reg2Inv & \textbf{100.0} & 57.7 & \textbf{94.8} & 80.4 & \underline{99.6} & 60.5 & \textbf{100.0} & 84.5 & 81.8 & \textbf{100.0} & 64.3 & 81.8 & 87.3 & 86.1 \\
        Simple3D & \textbf{100.0} & 74.8 & 70.3 & 59.6 & 95.4 & 82.4 & 87.1 & 81.2 & 86.4 & 96.2 & 89.5 & 85.5 & 81.5 & 86.0 \\
        CASL & \underline{98.1} & 83.2 & 76.1 & 56.3 & 83.8 & 89.0 & \underline{98.1} & \textbf{93.0} & 82.1 & 71.4 & \textbf{100.0} & \textbf{93.3} & 81.8 & 88.7 \\
        \midrule
        \multicolumn{15}{@{}l}{\textit{Unified}} \\
        MC3D-AD & 91.9 & \underline{84.1} & \underline{94.5} & \textbf{97.0} & 82.1 & 85.7 & 92.9 & 76.1 & 87.6 & \underline{97.6} & 93.8 & 67.0 & 73.6 & 84.2 \\
        SeDiR & 90.5 & \textbf{88.7} & 92.7 & 89.3 & 88.8 & \textbf{97.1} & 97.6 & 78.8 & \underline{92.4} & 94.8 & \textbf{100.0} & 87.3 & \textbf{94.2} & \underline{93.3} \\
        \textbf{\ours \ (Ours)} & \textbf{100.0} & 75.7 & 89.1 & \underline{93.3} & \textbf{100.0} & \underline{96.2} & \textbf{100.0} & \underline{85.8} & \textbf{93.3} & \textbf{100.0} & \textbf{100.0} & 89.4 & \underline{91.8} & \textbf{95.7} \\
        \bottomrule
    \end{tabular*}
\end{table*}

\begin{table*}[!htp]
    \centering
    \caption{Per-category point-level AUROC (\%) on Anomaly-ShapeNet. Bold and underlined values denote the best and second-best results, respectively.}
    \label{tab:shapenet_point}
    \fontsize{8}{9.0}\selectfont
    \setlength{\tabcolsep}{2.0pt}
    \renewcommand{\arraystretch}{0.95}
    \begin{tabular*}{\textwidth}{@{\extracolsep{\fill}}lrrrrrrrrrrrrrr@{}}
        \toprule
        Method & ashtray0 & bag0 & bottle0 & bottle1 & bottle3 & bowl0 & bowl1 & bowl2 & bowl3 & bowl4 & bowl5 & bucket0 & bucket1 & cap0 \\
        \midrule
        \multicolumn{15}{@{}l}{\textit{Category-specific}} \\
        BTF(Raw) & 51.2 & 43.0 & 55.1 & 49.1 & 72.0 & 52.4 & 46.4 & 42.6 & 68.5 & 56.3 & 51.7 & 61.7 & 68.6 & 52.4 \\
        BTF(FPFH) & 62.4 & 74.6 & 64.1 & 54.9 & 62.2 & 71.0 & 76.8 & 51.8 & 59.0 & 67.9 & 69.9 & 40.1 & 63.3 & 73.0 \\
        M3DM & 57.7 & 63.7 & 66.3 & 63.7 & 53.2 & 65.8 & 66.3 & 69.4 & 65.7 & 62.4 & 48.9 & 69.8 & 69.9 & 53.1 \\
        PatchCore(FPFH) & 59.7 & 57.4 & 65.4 & 68.7 & 51.2 & 52.4 & 53.1 & 62.5 & 32.7 & 72.0 & 35.8 & 45.9 & 57.1 & 47.2 \\
        PatchCore(PointMAE) & 49.5 & 67.4 & 55.3 & 60.6 & 65.3 & 52.7 & 52.4 & 51.5 & 58.1 & 50.1 & 56.2 & 58.6 & 57.4 & 54.4 \\
        CPMF & 61.5 & 65.5 & 52.1 & 57.1 & 43.5 & 74.5 & 48.8 & 63.5 & 64.1 & 68.3 & 68.4 & 48.6 & 60.1 & 60.1 \\
        Reg3D-AD & 69.8 & 71.5 & 88.6 & 69.6 & 52.5 & 77.5 & 61.5 & 59.3 & 65.4 & 80.0 & 69.1 & 61.9 & 75.2 & 63.2 \\
        IMRNet & 67.1 & 66.8 & 55.6 & 70.2 & 64.1 & 78.1 & 70.5 & 68.4 & 59.9 & 57.6 & 71.5 & 58.5 & 77.4 & 71.5 \\
        ISMP & 60.3 & 74.7 & 77.0 & 56.8 & 77.5 & 85.1 & 54.6 & 73.6 & 77.3 & 74.0 & 53.4 & 52.4 & 67.2 & 86.5 \\
        PO3AD & \underline{96.2} & 94.9 & 91.2 & 84.4 & 88.0 & 97.8 & 91.4 & 91.8 & 93.5 & \underline{96.7} & 94.1 & 75.5 & 89.9 & 95.7 \\
        PASDF & 91.9 & 95.8 & 95.1 & \underline{92.6} & \underline{94.8} & 96.3 & 90.0 & 81.6 & 93.9 & 86.5 & 90.9 & \textbf{87.5} & 82.4 & 94.8 \\
        Reg2Inv & 78.5 & \underline{99.1} & \underline{99.5} & 84.9 & 81.7 & 98.3 & 82.8 & 82.2 & 76.1 & 78.8 & 82.4 & 61.0 & 85.5 & 86.1 \\
        Simple3D & 92.0 & 95.4 & 97.4 & 72.8 & 83.8 & \underline{98.8} & 95.1 & 93.3 & \underline{99.3} & 92.9 & \underline{97.9} & 72.5 & 92.1 & \textbf{98.8} \\
        CASL & 88.7 & 96.8 & 85.3 & 86.2 & 91.4 & 95.8 & \textbf{97.5} & \underline{98.2} & \textbf{99.8} & \textbf{99.6} & 97.4 & 79.6 & \underline{92.7} & 97.4 \\
        \midrule
        \multicolumn{15}{@{}l}{\textit{Unified}} \\
        MC3D-AD & 80.1 & 81.5 & 89.5 & 88.3 & 90.1 & 82.7 & 53.4 & 60.7 & 78.3 & 65.4 & 55.2 & 79.0 & 89.5 & 86.7 \\
        SeDiR & 76.4 & 89.8 & 91.1 & 89.8 & 94.5 & 86.6 & 70.0 & 79.5 & 86.9 & 80.6 & 64.6 & 71.7 & 88.4 & 90.0 \\
        \textbf{\ours \ (Ours)} & \textbf{98.7} & \textbf{99.9} & \textbf{99.8} & \textbf{97.8} & \textbf{98.8} & \textbf{99.5} & \underline{97.2} & \textbf{99.7} & \textbf{99.8} & \textbf{99.6} & \textbf{99.1} & \underline{85.0} & \textbf{97.7} & \underline{98.0} \\
        \bottomrule
    \end{tabular*}
    \par\vspace{7pt}
    \begin{tabular*}{\textwidth}{@{\extracolsep{\fill}}lrrrrrrrrrrrrr@{}}
        \toprule
        Method & cap3 & cap4 & cap5 & cup0 & cup1 & eraser0 & headset0 & headset1 & helmet0 & helmet1 & helmet2 & helmet3 & jar \\
        \midrule
        \multicolumn{14}{@{}l}{\textit{Category-specific}} \\
        BTF(Raw) & 68.7 & 46.9 & 37.3 & 63.2 & 56.1 & 63.7 & 57.8 & 47.5 & 50.4 & 44.9 & 60.5 & 70.0 & 42.3 \\
        BTF(FPFH) & 65.8 & 52.4 & 58.6 & 79.0 & 61.9 & 71.9 & 62.0 & 59.1 & 57.5 & 74.9 & 64.3 & 72.4 & 42.7 \\
        M3DM & 60.5 & 71.8 & 65.5 & 71.5 & 55.6 & 71.0 & 58.1 & 58.5 & 59.9 & 42.7 & 62.3 & 65.5 & 54.1 \\
        PatchCore(FPFH) & 65.3 & 59.5 & 79.5 & 65.5 & 59.6 & 81.0 & 58.3 & 46.4 & 54.8 & 48.9 & 45.5 & 73.7 & 47.8 \\
        PatchCore(PointMAE) & 48.8 & 72.5 & 54.5 & 51.0 & 85.6 & 37.8 & 57.5 & 42.3 & 58.0 & 56.2 & 65.1 & 61.5 & 48.7 \\
        CPMF & 55.1 & 55.3 & 55.1 & 49.7 & 50.9 & 68.9 & 69.9 & 45.8 & 55.5 & 54.2 & 51.5 & 52.0 & 61.1 \\
        Reg3D-AD & 71.8 & 81.5 & 46.7 & 68.5 & 69.8 & 75.5 & 58.0 & 62.6 & 60.0 & 62.4 & 82.5 & 62.0 & 59.9 \\
        IMRNet & 70.6 & 75.3 & 74.2 & 64.3 & 68.8 & 54.8 & 70.5 & 47.6 & 59.8 & 60.4 & 64.4 & 66.3 & 76.5 \\
        ISMP & 73.4 & 75.3 & 67.8 & 86.9 & 60.0 & 70.6 & 58.0 & 70.2 & 68.3 & 62.2 & 84.4 & 72.2 & 82.3 \\
        PO3AD & 94.8 & 94.0 & 86.4 & 90.9 & 93.2 & 97.4 & 82.3 & 90.7 & 87.8 & \underline{94.8} & 93.2 & 84.6 & 87.1 \\
        PASDF & 86.1 & 89.4 & 92.0 & 94.8 & 88.4 & 94.5 & 86.3 & 89.1 & 81.6 & 64.6 & 80.9 & 95.8 & 95.9 \\
        Reg2Inv & 94.5 & 86.4 & \underline{97.0} & 79.8 & 88.1 & \underline{98.0} & \underline{94.6} & \underline{97.0} & \underline{92.5} & 90.6 & 89.1 & 95.6 & \underline{98.2} \\
        Simple3D & 96.4 & \textbf{97.9} & 96.4 & 97.9 & \underline{93.7} & 97.0 & 93.6 & 95.8 & 90.2 & 90.2 & \textbf{94.7} & 92.8 & 97.8 \\
        CASL & 97.1 & 94.9 & 66.9 & \underline{98.0} & 79.7 & 87.6 & 77.0 & 90.9 & 92.3 & 88.2 & 92.9 & \underline{98.0} & 97.4 \\
        \midrule
        \multicolumn{14}{@{}l}{\textit{Unified}} \\
        MC3D-AD & 92.2 & 88.0 & 88.5 & 82.9 & 71.5 & 85.6 & 66.6 & 70.9 & 74.4 & 58.2 & 81.7 & 60.2 & 84.8 \\
        SeDiR & \underline{97.2} & 94.0 & 95.0 & 87.1 & 84.2 & 74.8 & 70.8 & 68.7 & 81.0 & 59.6 & 88.5 & 70.1 & 91.1 \\
        \textbf{\ours \ (Ours)} & \textbf{98.5} & \underline{96.8} & \textbf{97.8} & \textbf{99.6} & \textbf{99.8} & \textbf{99.7} & \textbf{99.1} & \textbf{99.4} & \textbf{95.4} & \textbf{99.7} & \underline{93.7} & \textbf{99.5} & \textbf{99.8} \\
        \bottomrule
    \end{tabular*}
    \par\vspace{7pt}
    \begin{tabular*}{\textwidth}{@{\extracolsep{\fill}}lrrrrrrrrrrrrrr@{}}
        \toprule
        Method & phone & shelf0 & tap0 & tap1 & vase0 & vase1 & vase2 & vase3 & vase4 & vase5 & vase7 & vase8 & vase9 & Mean \\
        \midrule
        \multicolumn{15}{@{}l}{\textit{Category-specific}} \\
        BTF(Raw) & 58.3 & 46.4 & 52.7 & 56.4 & 61.8 & 54.9 & 40.3 & 60.2 & 61.3 & 58.5 & 57.8 & 55.0 & 56.4 & 55.0 \\
        BTF(FPFH) & 67.5 & 61.9 & 56.8 & 59.6 & 64.2 & 61.9 & 64.6 & 69.9 & 71.0 & 42.9 & 54.0 & 66.2 & 56.8 & 62.8 \\
        M3DM & 35.8 & 55.4 & 65.4 & 71.2 & 60.8 & 60.2 & 73.7 & 65.8 & 65.5 & 64.2 & 51.7 & 55.1 & 66.3 & 61.6 \\
        PatchCore(FPFH) & 48.8 & 61.3 & 73.3 & 76.8 & 65.5 & 45.3 & 72.1 & 43.0 & 50.5 & 44.7 & 69.3 & 57.5 & 66.3 & 58.0 \\
        PatchCore(PointMAE) & 88.6 & 54.3 & 85.8 & 54.1 & 67.7 & 55.1 & 74.2 & 46.5 & 52.3 & 57.2 & 65.1 & 36.4 & 42.3 & 57.7 \\
        CPMF & 54.5 & 78.3 & 45.8 & 65.7 & 45.8 & 48.6 & 58.2 & 58.2 & 51.4 & 65.1 & 50.4 & 52.9 & 54.5 & 57.3 \\
        Reg3D-AD & 59.9 & 68.8 & 58.9 & 74.1 & 54.8 & 60.2 & 40.5 & 51.1 & 75.5 & 62.4 & 88.1 & 81.1 & 69.4 & 66.8 \\
        IMRNet & 74.2 & 60.5 & 68.1 & 69.9 & 53.5 & 68.5 & 61.4 & 40.1 & 52.4 & 68.2 & 59.3 & 63.5 & 69.1 & 65.0 \\
        ISMP & 66.1 & 68.7 & 52.2 & 55.2 & 66.1 & 84.3 & 73.3 & 76.2 & 54.5 & 47.2 & 70.1 & 85.1 & 61.5 & 69.1 \\
        PO3AD & 81.0 & 66.3 & 78.3 & 69.2 & 95.5 & 88.2 & 97.8 & 88.4 & 90.2 & 93.7 & 98.2 & 95.0 & 95.2 & 89.9 \\
        PASDF & 95.1 & 86.5 & 88.4 & \underline{90.2} & 94.4 & 79.7 & 95.6 & 86.8 & 89.9 & 91.5 & 95.9 & 90.9 & 86.3 & 89.6 \\
        Reg2Inv & \underline{99.2} & 63.2 & \underline{91.8} & 86.9 & \underline{98.0} & 70.5 & \underline{99.7} & 84.4 & 92.7 & 87.9 & 86.3 & 93.4 & \underline{97.1} & 88.2 \\
        Simple3D & 96.4 & 90.1 & 85.7 & 81.1 & 93.4 & 80.7 & 98.6 & 91.0 & \underline{98.5} & \underline{96.6} & 99.0 & 97.5 & 91.3 & \underline{92.9} \\
        CASL & 83.8 & \underline{92.8} & 65.2 & 60.2 & 89.4 & \underline{97.2} & 97.1 & \underline{93.1} & 91.3 & 73.2 & \underline{99.7} & \textbf{98.9} & 88.6 & 89.9 \\
        \midrule
        \multicolumn{15}{@{}l}{\textit{Unified}} \\
        MC3D-AD & 85.4 & 65.8 & 52.4 & 53.7 & 87.0 & 72.6 & 82.2 & 79.7 & 78.6 & 64.7 & 61.6 & 88.5 & 77.6 & 75.9 \\
        SeDiR & 89.1 & 67.9 & 59.7 & 63.0 & 84.6 & 72.2 & 85.0 & 83.0 & 88.4 & 76.6 & 77.7 & 90.0 & 83.5 & 81.1 \\
        \textbf{\ours \ (Ours)} & \textbf{99.7} & \textbf{93.0} & \textbf{98.6} & \textbf{97.9} & \textbf{99.5} & \textbf{97.9} & \textbf{99.9} & \textbf{97.5} & \textbf{99.1} & \textbf{99.8} & \textbf{99.9} & \underline{97.6} & \textbf{97.6} & \textbf{98.2} \\
        \bottomrule
    \end{tabular*}
\end{table*}